%% file: root.tex
\documentclass[letterpaper,conference,10pt]{IEEEconf}
\IEEEoverridecommandlockouts
\usepackage{cite}

\usepackage{amsmath,amssymb,amsfonts}
\usepackage{algorithmic}
\usepackage{graphicx}
\usepackage{caption, subcaption}
\usepackage[dvipsnames]{xcolor}
\usepackage{mathtools}
\usepackage{tabularx}
\usepackage{statmath}
\usepackage{url}
\usepackage{svg}
\usepackage{tikz}
\usetikzlibrary{shapes,arrows}

\usetikzlibrary{arrows.meta,fit}

\tikzset{block/.style={draw, fill=blue!10, rectangle, thick,
minimum height=3em, minimum width=2em},
sum/.style={draw, fill=white, circle, thick, node distance=1cm},
input/.style={coordinate},
output/.style={coordinate},
pinstyle/.style={pin edge={latex[],thick,black}},
arrow/.style={draw,thick,-{latex[]}},
line/.style={draw,thick,-},
triangle/.style={draw,fill=red!20,regular polygon,regular polygon sides=3}}

\def\eqref#1{\textcolor{blue}{(\ref{#1})}}

\def\BibTeX{{\rm B\kern-.05em{\sc i\kern-.025em b}\kern-.08em
    T\kern-.1667em\lower.7ex\hbox{E}\kern-.125emX}}

\usepackage{graphicx} 
\graphicspath{{figures_gray/}} 

\makeatletter
\let\NAT@parse\undefined
\makeatother

\usepackage{hyperref}
\hypersetup{
    colorlinks,
    linkcolor=blue,
    urlcolor=blue,
    citecolor=blue,
    pdftitle={Motion Accuracy and Computational Effort in QP-based Robot Control},
}
\def\equationautorefname~#1\null{Equation (#1)\null}

\usepackage{booktabs}
\usepackage{algorithm2e}
\SetKwComment{Comment}{/* }{ */}

\SetAlgorithmName{Algorithm}{}{}

\usepackage{multirow}

\newcommand{\BIN}{\begin{bmatrix}}
\newcommand{\BOUT}{\end{bmatrix}}

\newcommand{\SO}[1]{\mbox{SO}(#1)}
\newcommand{\SE}[1]{\mbox{SE}(#1)}

\usepackage{siunitx}
\usepackage{blindtext}
\usepackage{tablefootnote}
\usepackage{stackengine}
    
\begin{document}

\title{\LARGE \bf Motion Accuracy and Computational Effort\\ in QP-based Robot Control}

\author{Sélim~Chefchaouni$^{1,*}$, Mehdi~Benallegue$^2$, Adrien~Escande$^1$ and~Pierre-Brice~Wieber$^1$
\thanks{This work was supported by the French government under the management of Agence Nationale de la Recherche (ANR) through the INEXACT project (ANR-22-CE33-0007), and the European Union’s Horizon 2020 research and innovation programme under the Marie Sklodowska-Curie grant agreement (grant no. 101066915).}%
\thanks{$^1$Inria Centre at Grenoble-Alpes University, Montbonnot-Saint-Martin, France.}%
\thanks{$^2$CNRS-AIST Joint Robotics Laboratory (JRL), Tsukuba, Japan.}%
\thanks{$^*$Corresponding author. E-mail: \href{mailto:selim.chefchaouni-moussaoui@inria.fr}{{selim.chefchaouni-moussaoui@inria.fr}}.}
}

\maketitle
\thispagestyle{plain}
\pagestyle{plain}

\begin{abstract}
Quadratic Programs (QPs) have become a mature technology for the control of robots of all kinds, including humanoid robots. One aspect has been largely overlooked, however, which is the accuracy with which these QPs should be solved. QP solvers aim at providing solutions accurate up to floating point precision ($\approx10^{-8}$). Considering physical quantities expressed in SI or similar units (meters, radians, \emph{etc}.), such precision seems completely unrelated to both task requirements and hardware capacity. Typically, humanoid robots never achieve, nor are capable of achieving sub-millimeter precision in manipulation tasks. With this observation in mind, our objectives in this paper are two-fold: first examine how the QP solution accuracy impacts the resulting robot motion accuracy, then evaluate how a reduced solution accuracy requirement can be leveraged to reduce the corresponding computational effort. Experiments with    a dynamic simulation of RHPS-1 humanoid robot indicate that computational effort can be divided by more than 27 while maintaining the desired motion accuracy.
\end{abstract}

\section{Introduction}
\label{sec:intro}

Quadratic Programs (QPs) have become a mature technology for the control of robots of all kinds, applied successfully to complex problems such as biped and quadruped locomotion~\cite{herdt:ar:2010} \cite{fahmi:ral:2019}, whole-body humanoid motion~\cite{cisneros_robust_2018} \cite{aircraft}, aerial manipulation~\cite{aerial}, \emph{etc}. They allow handling multiple simultaneous tasks while satisfying various physical constraints, but this usually comes at the cost of solving an optimization problem at each sampling period.

The corresponding computational effort incurs significant hardware requirements (and costs) for the embedded CPUs, and can also represent more than half of the energy consumption of the whole robot, even for a robot as complex as a humanoid~\cite{ECDP}~\cite{vision}~\cite{dlr}. Despite recent energy-efficient mechatronic designs~\cite{compdesign} \cite{balderashill:hal-03457432} and control approaches~\cite{dlr}, the typical energy autonomy of current humanoid robots doesn't exceed a few hours at most~\cite{1570316}~\cite{4755995}, which isn't satisfying for the applications usually targeted. Reducing the computational effort necessary to solve QPs in robot control laws could make a difference in this regard.

Solving a QP usually requires a series of matrix computations with algorithmic complexity $\mathcal{O}(n^3)$, where the dimension $n$ of the QP typically varies from $10$ to $1000$ in the case of humanoid robot control. These computations usually need to be updated at each sampling period, with typical control frequency $f$ from $100$ to $1000$~$[\SI{}{\hertz}]$~\cite{acmref}. The specific choice of which solver is used naturally impacts the computational effort to solve a QP~\cite{proxqp}. Also, it has already been shown that lower-frequency control laws ($5\sim15$~$[\SI{}{\hertz}]$) can be perfectly reasonable for biped balancing~\cite{villa_sensitivity_2019}, and that reducing the frequency of QP matrices updates can improve the performances of a controller in the case of predictive control schemes~\cite{mpc}. One aspect has been largely overlooked, however, which is the accuracy with which these QPs should be solved.

\newlength{\twosubht}
\newsavebox{\twosubbox}

\begin{figure}
    \centering
      \includegraphics[width=\linewidth]{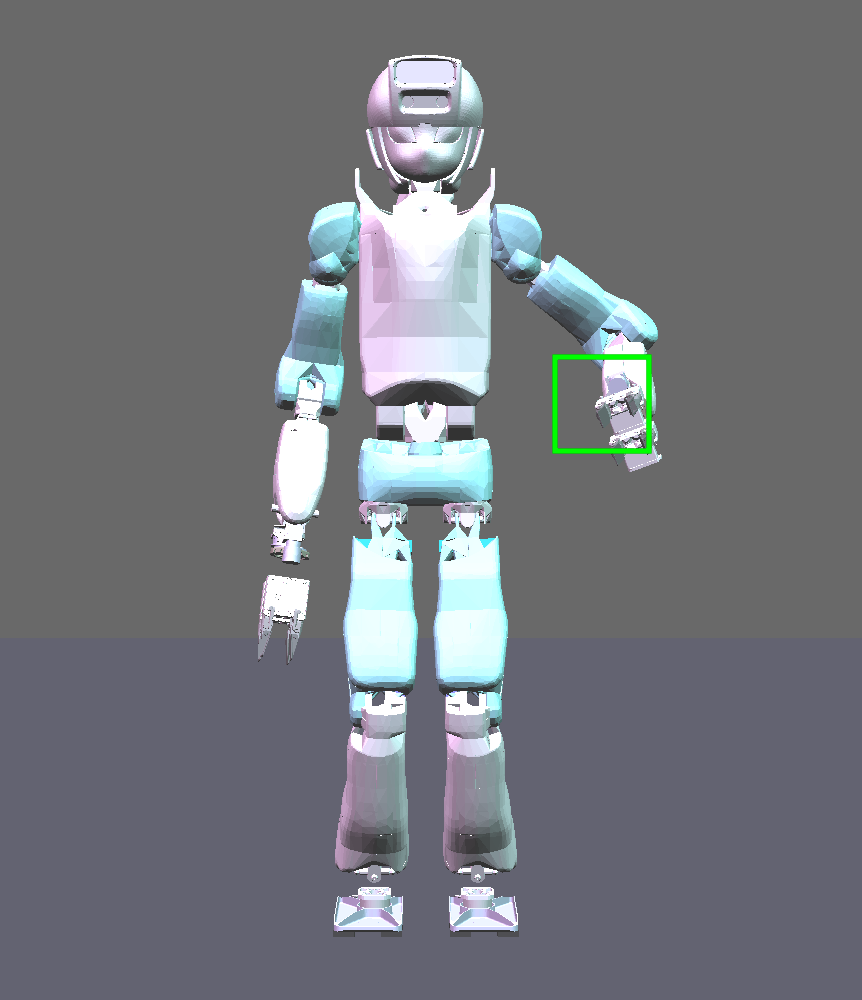}%
    \caption{Snapshot of simulated RHPS-1 robot controlled with a QP-based whole-body motion control law. The green square represents the desired left hand trajectory.}
  \label{fig:intro}
\end{figure}

Typical QP solvers aim at providing solutions accurate up to floating point precision ($\approx10^{-8}$). Considering physical quantities expressed in SI or similar units (meters, radians, \emph{etc}.), such precision is unrelated to both task requirements and hardware capacity. Typically, humanoid robots never achieve, nor are capable of achieving sub-millimeter precision in manipulation tasks. With this observation in mind, our objectives in this paper are two-fold: first examine how the QP solution accuracy impacts the resulting robot motion accuracy, then evaluate how a reduced solution accuracy requirement can be leveraged to reduce the corresponding computational effort.

To do so, we introduce in Section~\ref{sec:problem} a typical QP-based whole-body motion controller. We examine in Section~\ref{sec:sensitivity} how the QP solution accuracy impacts the resulting robot motion accuracy in a dynamic simulation of a RHPS-1  robot. We evaluate then in Section~\ref{sec:control} how a reduced solution accuracy requirement can be leveraged to reduce the corresponding computational effort, before concluding with a summary of findings and perspectives in Section~\ref{sec:conclusion}.

\section{Problem Formulation}
\label{sec:problem}

\subsection{System Representation}
\label{sec:repr}

We consider an articulated robot with a floating base and $n_j \in \mathbb{N}$ actuated joints. We define the robot's configuration, velocity and acceleration as
\begin{align}
\hat{q} &\in \SE3 \times \mathbb{R}^{n_j}\\
\dot{q} &\in \mathbb{R}^{6 + n_j}\\
\ddot{q} &\in \mathbb{R}^{6 + n_j}
\end{align}
where the configuration $\hat{q}$ involves an element of $\SE3$, the corresponding velocity being represented as an element of $\mathbb{R}^6$. As a result, the velocity $\dot{q}$ is not exactly the time-derivative of $\hat{q}$.

\begin{figure}
    \centering
    \vspace{1mm}
    \begin{subfigure}[b]{0.4\textwidth}
      \centerline{\begin{tikzpicture}[auto, node distance=2cm,scale=1.0, every node/.style={scale=1}]
        \input{tikz1}
      \end{tikzpicture}}
      \caption{}
      \label{fig:control_scheme1}
    \end{subfigure}
    \begin{subfigure}[b]{0.4\textwidth}
      \centerline{\begin{tikzpicture}[auto, node distance=2cm,scale=1.0, every node/.style={scale=1}]
        \input{tikz2}
      \end{tikzpicture}}
      \caption{}
      \label{fig:control_scheme2}
    \end{subfigure}
    \caption{(a) Control scheme for a position and speed controlled robot based on an acceleration obtained from a QP. (b) Same scheme with noise $\sigma$ added to the QP solution to model solution inaccuracy.}
    \label{fig:control_scheme}
\end{figure}
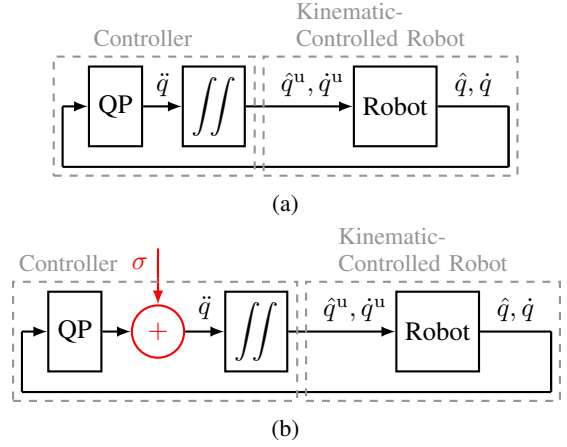

We consider a QP-based whole-body motion controller to compute at each sampling period an optimal control parameter
\begin{align}
x= \begin{bmatrix} \ddot{q} & \tau & f_c \end{bmatrix}^\top
\end{align}
including joint torques ${\tau} \in \mathbb{R}^{n_j}$~\cite{djeha_robust_2023} and $n_c$ contact forces $f_c \in \mathbb{R}^{3\times n_c}$ exerted by the robot feet on the ground~\cite{cisneros_robust_2018}. We consider, however, that this motion controller eventually provides a reference configuration $\hat{q}^\mathrm{u}$ and velocity $\dot{q}^\mathrm{u}$ to low-level motor controllers, as represented in Fig.~\ref{fig:control_scheme1}.

\subsection{Task-Based Approach}
\label{sec:tasks}

We define kinematic tasks and their derivatives as
\begin{align}
e &= \hat{g}(\hat{q}) \ominus \hat{g}^*\\ 
\dot{e} &= J(\hat{q})\dot{q} - \dot{g}^* \label{eq:taskError1}\\ 
\ddot{e} &= J(\hat{q})\ddot{q} + \dot{J}(\hat{q},\dot{q}) \dot{q} - \ddot{g}^* \label{eq:taskError2} 
\end{align}
using a difference operator $\ominus$ between a function of the robot configuration $\hat{g}(\hat{q})$ and a (possibly time-dependent) target value $\hat{g}^*$. This operator depends on the image space of the function: typically, $x-y$ over $\mathbb{R}^n$ and $\log(y^{-1}x)$ over $\SE3$ or $\SO3$. Note that $\ddot{e}$ is related to $\ddot{q}$ linearly through the Jacobian matrix $J(\hat{q})$.

In order to regulate these tasks and derivatives to 0, we consider a critically damped Proportional-Derivative (PD) target behavior with gain $k$:
\begin{equation}
    \ddot{e}^* = k\, e + 2\sqrt{k}\, \dot{e}.
\end{equation}
Our QP-based whole-body motion controller will then include constraints $\ddot{e} = \ddot{e}^*$ and objectives $\left\| \ddot{e} - \ddot{e}^*\right\|^2$ which are respectively linear and quadratic functions of $\ddot{q}$.

\subsection{Control Constraints}
\label{sec:const}

\subsubsection{Equations of Motion}
The robot must satisfy the equations of motion (EoM)
\begin{equation}
{M}(\hat{q})\, {\ddot{q}} + {h}(\hat{q}, \dot{q}) = {S}\, {\tau} + {{J}_{c}}(\hat{q})^\top {f}_{c}
\label{eq:newton_second_law}
\end{equation}
with an inertia matrix ${M}(\hat{q})$, inertial effects ${h}(\hat{q}, \dot{q})$, a selection matrix ${S} = [{0} \ {I}_{n_j}]^\top$ for joint actuators, and a contact Jacobian matrix ${J_{c}(\hat{q})}$.

\subsubsection{Feet Motion} To keep both feet on the ground, we regulate their position and orientation:
\begin{subequations}\label{eq:feet_pos}
\begin{align}
e_{\mathrm{lf}} &= \xi_{\mathrm{lf}}(\hat{q}) \ominus \xi_{\mathrm{lf}}^* \\
e_{\mathrm{rf}} &= \xi_{\mathrm{rf}}(\hat{q}) \ominus \xi_{\mathrm{rf}}^*.
\end{align}
\end{subequations}
where $\xi_{\mathrm{lf}}$ and $\xi_{\mathrm{rf}}$ refer to the left and right foot respectively.

\subsubsection{Friction Cones} Contact forces must stay inside friction cones, which we approximate as polytopes~\cite{bouyarmane2018}:
\begin{equation}
f_c \in \mathcal{C}.
\label{eq:cones}
\end{equation}

\subsubsection{Joint Limits} Joint motion must respect acceleration limits (that include position and velocity limits through a velocity damper~\cite{vaillant_multi-contact_2016}) and torque limits:
\begin{align}
\ddot{q}^- &\leq \ddot{q} \leq \ddot{q}^+\label{eq:joint_limits_2}\\
\tau^- &\leq \tau \leq \tau^+.\label{eq:joint_limits_1}
\end{align}

\subsection{Control Objectives}
\label{sec:objectives}

\subsubsection{Left Hand Motion} In the experiments that follow, the main task for the robot is to track a desired trajectory (a vertical square with fixed hand orientation) with a frame attached to the left hand of the robot:
\begin{equation}
e_{1} =\xi_{\mathrm{lh}}(\hat{q}) \ominus \xi_{\mathrm{lh}}^*(t).
\label{eq:pose_task}
\end{equation}

\subsubsection{Center of Mass} The Center of Mass (CoM) of the robot must stay nevertheless at a fixed position above the support polygon to preserve balance:
\begin{equation}
\label{eq:com_task}
e_{2} = x_\mathrm{CoM}(\hat{q}) \ominus x_\mathrm{CoM}^*.
\end{equation}

\subsubsection{Body Posture} A desired configuration $\hat{q}^*$ is specified:
\begin{equation}
e_{3} = P (\hat{q} \ominus \hat{q}^*)
\label{eq:posture_task}
\end{equation}
where $P$ is a diagonal weight matrix with lower weights on the left arm.

\subsubsection{QP convexity} To ensure the uniqueness of the solution, we include an objective involving contact forces, making the QP strictly convex:
%
\begin{equation}
e_4 = f_c - f_c^*.
\label{eq:damping_task}
\end{equation}

\subsection{QP Formulation}
\label{sec:formulation}

Following the procedure outlined in Section~\ref{sec:tasks}, these control constraints \eqref{eq:newton_second_law}-\eqref{eq:joint_limits_1} and objectives \eqref{eq:pose_task}-\eqref{eq:damping_task} can be summarized in the following QP, using weights $w_i$:
\begin{equation}
    \begin{split}\begin{array}{ll}
    \underset{x}{\mbox{minimize}} &
    \displaystyle \sum_{i=1}^3 w_i \left\|\ddot{e}_i - \ddot{e}_i^*\right\|^2 + w_4 \left\|e_4\right\|^2 \\
    \mbox{subject to} & \mbox{constraints derived from \eqref{eq:newton_second_law} to \eqref{eq:joint_limits_1}},
    \end{array}\end{split}
    \label{eq:qp_formulation}
\end{equation}
or in canonical form,
\begin{equation}
\begin{split}\begin{array}{ll}
\underset{x}{\mbox{minimize}} &
\frac{1}{2} \| R x - s \|^2_W = \frac{1}{2} x^\top G\, x  +    a^\top x + b\\
\mbox{subject to} & l \leq C x \leq u
\end{array}\end{split}
\label{eq:qp_formulation_full}
\end{equation}
where $G = R^\top W R \in \mathbb{R}^{n\times n}$ is a symmetric positive definite Hessian matrix, $a = -R^\top W s \in \mathbb{R}^{n}$ is the linear part of the objective function, $b\in \mathbb{R}^n$ is a constant vector, $C \in \mathbb{R}^{n\times m}$ is the constraints matrix with $l \in \mathbb{R}^m$ and $u \in \mathbb{R}^m$ the corresponding bounds (we can have $l_i = u_i$ for some $i$ to define equality constraints).

In practice, by separating the joint dynamics (denoted with subscript $_l$) from the floating base dynamics, the torques $\tau$ can be removed from this QP and computed separately with
\begin{equation}
{\tau} = {M}_l(\hat{q})\, {\ddot{q}} + {h}_l(\hat{q}, \dot{q}) - {{J}_{c,l}}(\hat{q})^\top {f}_{c},
\label{eq:newton_second_law_replace_tau}
\end{equation}
then replaced in \eqref{eq:newton_second_law} and \eqref{eq:joint_limits_2} (as well as $\tau^-$ and $\tau^+$). This helps reduce the computational effort to solve the QP by reducing the number of variables and constraints involved~\cite{tau}. Consequently, we work with a reduced control parameter
\begin{align}
x= \begin{bmatrix} \ddot{q} & f_c \end{bmatrix}^\top
\end{align}
in the following. This QP is implemented in a closed-loop controller, as represented in Fig.~\ref{fig:control_scheme1}.

\section{Motion Accuracy as a Function of QP Solution Accuracy} 
\label{sec:sensitivity}

\subsection{Desired Motion Accuracy}

The desired accuracy for each control objective introduced in the previous Section can vary, depending on the task that the robot has to fulfill. For typical tasks such as nailing \cite{nailing} or peg-in-hole \cite{peginhole}, we consider that the left hand position accuracy
\begin{equation}
e_\mathrm{x} = \left\|x_\mathrm{lh} - x_\mathrm{lh}^*\right\|
\end{equation}
should ideally stay below $10^{-3}~[\SI{}{\meter}]$. Similarly, we consider that the hand orientation accuracy
\begin{equation}
e_\mathrm{r} = \left\| \log \left( R_\mathrm{lh}^\top\, R_\mathrm{lh}^* \right) \right\|
\end{equation}
should ideally stay below $10^{-3}~[\SI{}{\radian}]$. Since radians are defined with respect to arc length, we believe this provides a measure that can be directly compared to position accuracy.

Considering the typical size of the support polygon, we consider that the CoM position accuracy
\begin{equation}
e_\mathrm{CoM} = \left\| x_\mathrm{CoM} - x_\mathrm{CoM}^*\right\|
\end{equation}
should stay below $10^{-2}~[\SI{}{\meter}]$ to preserve balance of the robot.

Regarding the posture of the rest of the body, we propose to measure the joint centers' deviations in Cartesian space, as the impact of a deviation measured in joint space would deeply depend on the dimensions of the attached body. The deviation is measured as follows:
\begin{equation}
e_\mathrm{post} = \frac{1}{h}\sum_{j \in \mathcal{H}} \left\| x_j - x_j^* \right\|
\end{equation}
with $x_j$ the Cartesian position of joint $j$, $x_j^*$ its desired position obtained from the desired configuration $\hat{q}^*$, $\mathcal{H}$ the set of indices of joints not included in the left arm, which is discarded here, and $h$ the size of $\mathcal{H}$. We believe that a reasonable accuracy here would be also $10^{-2}~[\SI{}{\meter}]$ to avoid kinematic issues such as collisions. All these values are collected in Table~\ref{tab:x}. 

\begin{table}
    \begin{center}
    \vspace{2mm}
    \begin{tabular}{ll}
        \toprule
        Control objective & Desired accuracy  \\
        \midrule
        Left hand position & $10^{-3}~[\SI{}{\meter}]$ \\
        Left hand orientation & $10^{-3}~[\SI{}{\radian}]$ \\
        CoM position & $10^{-2}~[\SI{}{\meter}]$ \\
        Joint positions & $10^{-2}~[\SI{}{\meter}]$ \\
        \bottomrule
    \end{tabular}
    \end{center}
    \caption{Desired motion accuracy for each control objective of the robot.}
    \label{tab:x}
\end{table}

\begin{table}
    \centering
    \begin{tabular}{lll}
        \toprule
        Control objective & $k\,\left[\SI{}{\per\square\second}\right]$ & $w_i$ \\
        \midrule
        Left hand position & cf.\@ Table \ref{tab:my_label2} & 2000 \\
        Left hand orientation & cf.\@ Table \ref{tab:my_label2} & 5000 \\
        CoM position ($x$ and $y$ axes) & $100$  & $1000$ \\
        CoM position ($z$ axis)& $100$  & $200$ \\
        Body posture (left arm) & $50$  & $1$ \\
        Body posture (other joints) & $50$ & $5000$ \\
        Contact forces & -- & $0.01$ \\
        \bottomrule
    \end{tabular}
    \caption{Gains $k$ and weights $w_i$ for the different control objectives.}
    \label{tab:my_label}
\end{table}

\subsection{The Impact of QP Solution Accuracy}

\begin{figure*}
    \centering
        \includegraphics[scale=0.98]{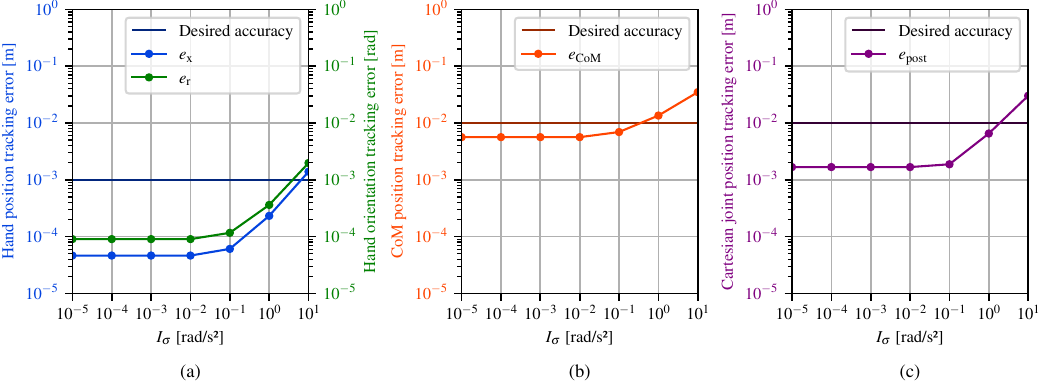}
    \caption{Tracking errors as functions of noise level $I_\sigma$ added to the QP solution: (a) hand position and orientation, (b) CoM position and (c) joint positions.} 
    \label{fig:task_errors}
\end{figure*}

We consider a dynamic simulation of a RHPS-1 humanoid robot~\cite{rhps1} with $n_j ={42}$  articulated joints, controlled with the QP~\eqref{eq:qp_formulation} (dimensions $n={72}$ and $m={124}$) using parameters from Table~\ref{tab:my_label} at a standard frequency $f=200~[\SI{}{\hertz}]$, implemented in Python using the Pinocchio library~\cite{carpentier2019pinocchio}. To model solution accuracy, we introduce a uniform noise
\begin{equation}
\sigma \sim \mathcal{U}(-I_\sigma, +I_\sigma)
\end{equation}
added to the solution of the QP with different values of $I_\sigma$, as shown in Fig.~\ref{fig:control_scheme2}. We can observe in Fig.~\ref{fig:task_errors} that this added uniform noise has little to no visible effect on motion accuracy for values of $I_\sigma$ as large as $10^{-2}$, and that the desired motion accuracy specified in Table~\ref{tab:x} is even satisfied for values as large as $10^0$ ($10^{-1}$ for the CoM position), indicating that it may be completely unnecessary to compute solutions of the QP~\eqref{eq:qp_formulation} more accurately than this value.

\section{Reduced Accuracy Allows Reduced Computational Effort}
\label{sec:control}

\subsection{Updating QP Matrices Less Frequently}
\label{sec:matupdate}

The QP~\eqref{eq:qp_formulation} is solved at each sampling period with updated vectors and matrices. In our experiments, we solve it using a C++ implementation of a standard Goldfarb-Idnani dual active-set method~\cite{goldfarb_numerically_1983}, \verb+jrl-qp+~\cite{jrlqp}. In this algorithm, expensive matrix computations with complexity $\mathcal{O}(n^3)$ and $\mathcal{O}(mn^2)$ are done once, at the beginning of the solution process:
\begin{itemize}
\item[-] a Cholesky decomposition of the Hessian matrix $G =LL^\top$,
\item[-] a matrix inversion $J = L^{-\top}$,
\item[-] a matrix product $B = J^\top N$ where $N$ is a selection of columns of $C^\top$,
\item[-] a QR decomposition of $B = QR$,
\item[-] and a matrix product $H = JQ$.
\end{itemize}
Updating these matrix computations only every $r$ sampling periods could save a lot of computational effort. This corresponds to keeping the matrices $G$ and $C$ of the QP~\eqref{eq:qp_formulation_full} constant over $r$ sampling periods while still updating its vectors $a$, $l$ and $u$. That would lead to potentially incorrect solutions, because these matrices vary with time, but we have seen that inaccurate solutions can have little to no visible effect on the resulting robot motion.

We can observe in Fig.~\ref{fig:err_fun_time} how this update ratio $r$ affects both the hand position and orientation tracking errors and the computational effort to solve the QP~\eqref{eq:qp_formulation}, measured over $1~\left[\SI{}{\second}\right]$ of robot motion using the C++ \verb+std::time+ library on an Intel\textsuperscript{®} Core™ i7-13800H CPU. For update ratios up to $r=22$, corresponding to a matrix update frequency
\begin{equation}
    f_\mathrm{u}=\frac{f}{r}\approx 9~\left[\SI{}{\hertz}\right],
\end{equation}
there are no visible effects on hand motion tracking error, whereas the computational effort is reduced by more than $15$ with respect to a standard controller (when $r=1$). Beyond this value, motion accuracy starts degrading steeply with little additional gain in computational effort, making this value an obvious optimal choice.

\subsection{Updating the QP Solution Less Frequently}

In addition to updating the QP matrices less frequently, we could also update the QP solution itself less frequently. This corresponds to keeping both the matrices $G$ and $C$ and the vectors $a$, $l$ and $u$ of the QP~\eqref{eq:qp_formulation_full} constant over a larger period of time. But this corresponds to lowering the control frequency, and this requires lowering the feedback gain accordingly in order to keep the control law stable, what contributes to degrading the tracking accuracy. The feedback gain $k$ that we used for the left hand task is chosen empirically to minimize tracking error, and can be found in Table~\ref{tab:my_label2} for varying control frequencies $f$. 

The update ratio $r$ of the QP matrices, and the corresponding update frequency $f_\mathrm{u}$ need to be adapted as well to the control frequency $f$, as specified in Table~\ref{tab:my_label2}. We specifically look for the highest ratio $r$ such that tracking errors don't increase more than $5~\SI{}{\percent}$ compared to the case when $r={1}$. We keep the feedback gains for other objectives fixed, however, as already stated in Table~\ref{tab:my_label}. Interestingly, the optimal choice for the matrices update frequency always stays around $f_\mathrm{u}\approx 9~[\SI{}{\hertz}]$.

\begin{figure*}
    \centering
    \includegraphics[scale=0.98]{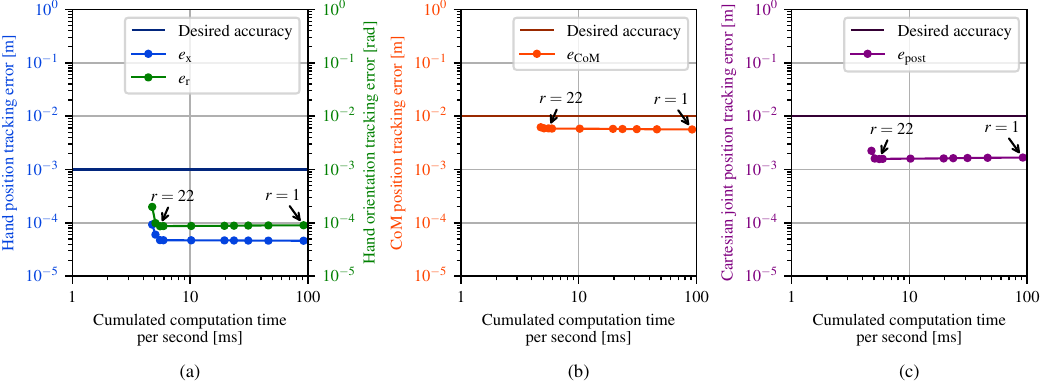}
    \caption{Tracking errors and computational effort over $1~\left[\SI{}{\second}\right]$ of robot motion for varying QP matrices update ratios $r$.
    }
    \label{fig:err_fun_time}
\end{figure*}

\begin{figure*}
    \centering
    \includegraphics[scale=0.98]{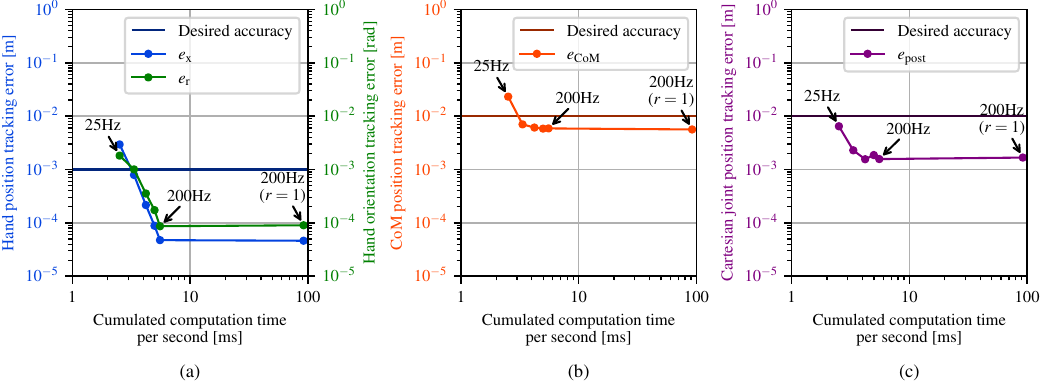}
    \caption{Tracking errors and computational effort over $1~\left[\SI{}{\second}\right]$ of robot motion for varying control frequencies $f$, feedback gains $k$ and matrices update ratios $r$ as specified in Table \ref{tab:my_label2}. An additional point with standard values ($f=200~[\SI{}{\hertz}]$, $r=1$) is added for comparison.}
    \label{fig:err_fun_time_varfreq}
\end{figure*}

\begin{figure}[t]
    \centering
    \includegraphics[scale=0.98]{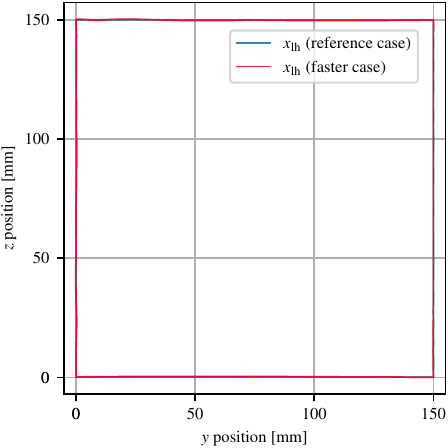}
    \caption{Left hand motion in the $yz$ plane. The left hand trajectory $x_\mathrm{lh}$ is plotted for the reference case ($f=200~[\SI{}{\hertz}]$, $r=1$) and the faster case satisfying the desired accuracy ($f=50~[\SI{}{\hertz}]$, $r=5$).}
    \label{fig:square_trajectory}
\end{figure}

We can observe in Fig.~\ref{fig:err_fun_time_varfreq} how this combination of varying control frequency, feedback gain and matrices update ratio affects both the hand position and orientation tracking errors and the computational effort to solve the QP~\eqref{eq:qp_formulation}, measured as before over 1~$\left[\SI{}{\second}\right]$ of robot motion. With respect to the results already presented in Fig.~\ref{fig:err_fun_time}, we can further reduce the computational effort by lowering the control frequency, but this comes at the cost of reducing motion accuracy. The desired accuracy specified in Table~\ref{tab:x} can still be satisfied, however, with a control frequency as low as $50~[\SI{}{\hertz}]$, allowing a total reduction of computational effort by more than $27$, from $91.9~[\SI{}{\milli\second}]$ per second of motion, down to $3.3~[\SI{}{\milli\second}]$. As displayed in Fig.~\ref{fig:square_trajectory}, the left hand motion obtained with these parameters has no visible difference with the motion obtained in the reference case.

\begin{table}
    \centering
    \vspace{1mm}
    \begin{tabular}{rrrr}
        \toprule
        Control freq. & Feedback gain & Update ratio & Update freq. \\
        $f$~$\left[\SI{}{\hertz}\right]$ & $k$~$\left[\SI{}{\per\square\second}\right]$ & $r$ & $f_\mathrm{u}$~$\left[\SI{}{\hertz}\right]$ \\
        \midrule
        $200$ & $18000$ & $22$ & $9.1$ \\
        $150$ & $10000$ & $16$ & $9.4$ \\
        $100$  & $4000$ & $12$ & $8.3$ \\
        $50$  & $1150$ & $5$ & $10$ \\
        $25$ & $280$ & $3$ & $8.3$ \\
        \bottomrule
    \end{tabular}
    \caption{Feedback gain $k$~$\left[\SI{}{\per\square\second}\right]$ for the left hand motion (position and orientation), QP matrices update ratio $r$ and corresponding frequency $f_\mathrm{u}$~$\left[\SI{}{\hertz}\right]$ for varying control frequencies $f$~$\left[\SI{}{\hertz}\right]$.} 
    \label{tab:my_label2}
\end{table}

\section{Conclusion}
\label{sec:conclusion}

In this paper, we considered a dynamic simulation of a RHPS-1 humanoid robot standing on both legs while tracking a trajectory with its left hand, using a standard QP-based whole-body motion control law. We examined how the QP solution accuracy impacts the resulting robot motion accuracy, and evaluated how this can be leveraged to reduce the corresponding computational effort. The result is that we can divide this computational effort by more than $27$ while maintaining the desired motion accuracy. This is obtained by carefully choosing the frequency at which the matrices $G$ and $C$ and the vectors $a$, $l$ and $u$ of the QP~\eqref{eq:qp_formulation_full} are updated. The key observation is that sufficient motion accuracy is perfectly possible with very degraded QP solutions, which can be obtained in turn with heavily reduced computational effort. These initial findings need to be confirmed now in more complex, more dynamic scenarios (\textit{e.g.} walking), on real hardware. We believe, however, that this shows a promising direction to reduce the computational burden of QP-based robot control laws.

\bibliographystyle{IEEEtran.bst} 
\bibliography{references.bib} 

\end{document}

%% file: tikz1.tex
\node [input](input){};
\node [block, fill=white, right of=input, node distance=0.7cm](fbcont){QP};
\node [block, fill=white, right of=fbcont, node distance=1.3cm](act){$\displaystyle\iint$};
\node [fill=white, right of=act, node distance=0.9cm](actout){};
\node [block, fill=white, right of=actout, node distance=1.5cm](plant){Robot};
\node [output, right of=plant](out1){};
\node [output, below of=out1](out2){};
\node [input, right of=plant, node distance=1.5cm](interim1){};
\node [input, below of=interim1, node distance=0.8cm](interim2){};
\node [input, below of=input, node distance=0.8cm](interim3){};


\draw [arrow] (fbcont)--node{$\ddot{q}$}(act);
\draw [arrow] (act)--node{$\quad \hat{q}^\mathrm{u}, \dot{q}^\mathrm{u}$}(plant);
\draw [line] (plant)--node{$\hat{q}, \dot{q}$}(interim1);
\draw [line] (interim1)--(interim2);
\draw [line] (interim1)--(interim2);
\draw [line] (interim2)--(interim3);
\draw [line] (interim3)--(input);
\draw [arrow] (input)--(fbcont);

\node[fit=(interim3)(fbcont)(act)(input),draw,Gray,dashed, thick, label={[Gray]above:{{\parbox{1.6cm}{\small Controller}}}}]{};
\node[fit=(interim1|-interim2)(plant)(actout),draw,Gray,dashed, thick, label={[Gray]above:{{\parbox{2.5cm}{\small Kinematic- Controlled Robot}}}}]{};

%% file: tikz2.tex
\node [input](input){};
\node [block, fill=white, right of=input, node distance=0.7cm](fbcont){QP};
\node [sum, right of=fbcont, node distance=1.1cm, color=red, fill=white](sum2){$+$};
\node [block, fill=white, right of=sum2, node distance=1.3cm](act){$\displaystyle\iint$};
\node [fill=white, right of=act, node distance=0.9cm](actout){};
\node [block, fill=white, right of=actout, node distance=1.5cm](plant){Robot};
\node [output, right of=plant](out1){};
\node [output, below of=out1](out2){};
\node [input, right of=plant, node distance=1.5cm](interim1){};
\node [input, below of=interim1, node distance=0.8cm](interim2){};
\node [input, below of=input, node distance=0.8cm](interim3){};
\node [input, above of=sum2, node distance=0.7cm](noise){};
\node [input, above of=noise, node distance=0.4cm](noiseup){};


\draw [arrow] (fbcont)--node{}(sum2);
\draw [arrow] (sum2)--node{$\ddot{q}\vphantom{\ddot{q}_\mathrm{QP}}$}(act);
\draw [arrow] (act)--node{$\quad \hat{q}^\mathrm{u}, \dot{q}^\mathrm{u}$}(plant);
\draw [line] (plant)--node{$\hat{q}, \dot{q}$}(interim1);
\draw [line] (interim1)--(interim2);
\draw [line] (interim1)--(interim2);
\draw [line] (interim2)--(interim3);
\draw [line] (interim3)--(input);
\draw [arrow] (input)--(fbcont);
\draw [line, color=red] (noise)--node[color=red]{$\sigma$}(noiseup);
\draw [arrow, color=red] (noise)--(sum2);

\node[fit=(sum2|-interim3)(fbcont)(act)(input),draw,Gray,dashed, thick, label={[Gray]above:{{\parbox{3.6cm}{\small Controller}}}}]{};
\node[fit=(interim1|-interim2)(plant)(actout),draw,Gray,dashed, thick, label={[Gray]above:{{\parbox{2.5cm}{\small Kinematic- Controlled Robot}}}}]{};